\documentclass[conference]{IEEEtran}
\IEEEoverridecommandlockouts
\usepackage{cite}
\usepackage{amsmath,amssymb,amsfonts}
\usepackage{algorithmic}
\usepackage{graphicx}
\usepackage{textcomp}
\usepackage{float}
\usepackage{amsmath}
\usepackage{subcaption}
\usepackage{booktabs}
\usepackage{multirow}
\usepackage{array}
\usepackage{longtable}
\usepackage{lipsum}
\usepackage{xcolor}

\def\BibTeX{{\rm B\kern-.05em{\sc i\kern-.025em b}\kern-.08em
    T\kern-.1667em\lower.7ex\hbox{E}\kern-.125emX}}
\begin{document}
\twocolumn[
\begin{center}
    
    
    \large
    \textbf{MaskUno: Switch-Split Block For Enhancing Instance Segmentation} \\
    \LARGE
    
    \vspace*{0.5 cm}

\small
\textbf{Jawad Haidar, Marc Mouawad, Imad Elhajj, Daniel Asmar}
\vspace{0.2 cm}
\noindent

\raggedright
\footnotesize
\textit{Vision and Robotics Lab (VRL), Maroun Semaan Faculty of Engineering and Architecture, American University of Beirut, Beirut, Lebanon
}

\vspace*{1. cm}
\end{center}]

\begin{abstract}
Instance segmentation is an advanced form of image segmentation which, beyond traditional segmentation, requires identifying individual instances of repeating objects in a scene. Mask R-CNN is the most common architecture for instance segmentation, and improvements to this architecture include steps such as benefiting from bounding box refinements, adding semantics, or backbone enhancements. In all the proposed variations to date, the problem of competing kernels (each class aims to maximize its own accuracy) persists when models try to synchronously learn numerous classes. In this paper, we propose mitigating this problem by replacing mask prediction with a Switch-Split block that processes refined ROIs, classifies them, and assigns them to specialized mask predictors. We name the method MaskUno and test it on various models from the literature, which are then trained on multiple classes using the benchmark COCO dataset. An increase in the mean Average Precision (mAP) of 2.03\% was observed for the high-performing DetectoRS when trained on 80 classes. MaskUno proved to enhance the mAP of instance segmentation models regardless of the number and type of classes or the type of architecture in which it was applied.

\end{abstract}


\section{Introduction} 
Instance segmentation is a fundamental task in computer vision, which combines mask segmentation and object detection, and whose goal is to perform a per-pixel classification of different objects in a scene. Instance segmentation has a wide range of applications such as in medical diagnostics \cite{xu2016gland}, agricultural analysis \cite{minervini2016finely}, and autonomous driving \cite{geiger2012we}. Its abundant use in practical scenarios necessitates accurate and robust predictions. The problem of instance segmentation is difficult and faces challenges such as occlusion, scale changes, and cluttered backgrounds  \cite{gu2022review}.   

With an eye toward enhancing the performance of instance segmentation, some researchers adopted the cascade methods that rely on progressive refinement \cite{cai2018cascade}. For example, one could pipe the output of the the bounding box prediction from a Stage 1 cascade into a region extractor (Stage 2), and so on, such that the progressive flow of the refined bounding boxes guarantees better-localized feature extraction. As a result, a higher accuracy is achieved in delineating masks and bounding boxes. 
    
Another research direction concentrates on designing backbone networks that can detect features that are better suited for the downstream task of instance segmentation. There exists a trade-off between the depth of a backbone model and the spatial resolution of features. Contrary to image classification, instance segmentation requires high spatial resolution of features, and so research on this front is concerned in architecting backbone networks that can identify features with high spatial resolutions
     \cite{qiao2021detectors,du2020spinenet}.  

Most previous work in instance segmentation adopts the pipeline of Mask R-CNN \cite{cai2018cascade,du2020spinenet,qiao2021detectors,lin2017focal,chen2019hybrid}: after extracting features using a backbone, a Region Proposal Network (RPN) localizes the regions that encompass the sought-after objects. The localized regions are then extracted using a region of interest block and fed to detection and mask generation branches. To avoid competition among classes, Mask R-CNN replaces the multi-nomial cross entropy with a per-class sigmoid and binary loss \cite{he2017mask}, keeping in mind that the weights are common between different classes given that they belong to the same branch.      
To tackle this gap, our method works on top of most instance segmentation methods by simply isolating the branches corresponding to different classes. Specifically, our pipeline consists of two main parts: detection and per-class segmentation. The output of the detection is fed to a switch that directs the bounding boxes to their corresponding segmentation branch. Consequently, the optimization of each branch is done independently to guarantee non-competing kernels between different classes. These simple modifications significantly improve the performance of the baseline Mask R-CNN \cite{he2017mask} and the high performing CNN based model Detectors \cite{qiao2021detectors}.

The remainder of this paper is organized as follows. In the next section, we will delve deeper into the recent methods in the field of instance segmentation. In Section \ref{sec:methodology}, the MaskUno proposed method is introduced. In Section \ref{sec:Experimentation and Discussion}, experiments are conducted and results demonstrate the enhancements obtained by MaskUno when applied to various instance segmentation models, trained and tested on the standard COCO dataset benchmark. Finally, Section \ref{sec: conclusion} concludes the paper.
\vspace{-2.22mm}
\section{Related Work}
\label{sec:relatedwork}
The architecture of an instance segmentation model contains different blocks, each contributing in a different manner to the end result. First, the backbone is responsible for extracting high-level features of the image. One can usually add a Feature Pyramid Network (FPN) that can help the network gain multi-scale context, which is crucial for detecting objects of different sizes. Next, comes the RPN, which generates from a feature map potential region candidates that might contain objects. The ROI-Align layer ensures that all the regions of interest are of similar size. Finally, the regions of interest are fed to a terminal block which classifies them and predicts bounding boxes and binary masks for each instance. In the following section, we will review the various interventions at the level of the backbone, object detection, and cascade. 

\subsection{Backbones} 
The pillar of every deep learning model is the feature extraction block (named backbone) and most of them are built around a CNN. The downside of such models is that spatial resolution decreases with depth, which is problematic for instance segmentation models that require high spatial resolution for pixel-wise mask prediction. To address this issue, SpineNet \cite{du2020spinenet} uses Neural Architecture Search to learn scale permuted features as well as cross-scale connections. DetectoRS \cite{qiao2021detectors} proposes a recursive FPN  that mimics thinking twice at a macro level; more specifically, the outputs of the FPN are fed to the bottom-up backbone. Additionally, they introduce the looking twice concept at the micro level, which is achieved by adding Switchable Atrous Convolutions.

\subsection{Object detection} 
Object detection is an essential block for instance segmentation as it delineates the bounding box surrounding each of the detected objects in a scene. 
Object detection methods can be based either on one stage\cite{redmon2016you,liu2016ssd,lin2017focal,law2018cornernet,duan2019centernet,zhou2019bottom}, or two stages \cite{girshick2014rich,he2017mask,dai2016r,pang2019libra}, where the former achieves better inference speed and the latter provides higher accuracy. The high inference speed in one-stage detectors is ascribed to their simple architecture; for instance, YOLO \cite{redmon2016you} divides the image into a grid and places various anchors having different scales and aspect ratios. Then, the anchors are directly classified and adjusted after a feature extraction stage. For the sake of predicting small-scale objects, SSD \cite{liu2016ssd} performs detection at different stages of the feature extraction. RetinaNet \cite{lin2017focal} proposes the focal loss to attenuate the effect of the abundant easy negatives, resulting in improved accuracy over the standard cross entropy loss. To get rid of anchors, CornerNet \cite{law2018cornernet}  expresses the bounding boxes by two points, the top-left and bottom-right corners. CenterNet \cite{duan2019centernet} uses the corners as proposals, and the center to verify the class of the object. A more accurate object representation is proposed by ExtremeNet \cite{zhou2019bottom} that replaces the corners with the objects' extreme points.

On the other hand, two-stage object detectors include an RPN, whose role is to first propose regions with high objectiveness probability, and output class-agnostic bounding boxes of possible objects. These bounding boxes are then used by the region of interest extractor.

R-CNN \cite{girshick2014rich} uses a non-learning-based proposal generator to propose possible objects, and then each of these regions of interest is passed through a CNN to extract corresponding features. This is a slow operation as each box is passed independently. On the other hand, Fast R-CNN \cite{girshick2015fast} applies a feature extractor to the entire image and then extracts the corresponding features. Finally, Faster R-CNN \cite{ren2015faster}, replaces the non-learning-based proposal method with a learnable RPN. Mask R-CNN \cite{he2017mask} adds a mask prediction head to support instance segmentation, and additionally proposes a region of interest alignment block that takes into account the neighboring pixels surrounding the bounding box.

\subsection{Cascade} 
Cascade is the process of iterating the output of a model to achieve progressive refinement. Cascade R-CNN \cite{cai2018cascade} proposes to add a simple cascade architecture on top of Faster R-CNN. A Stage 1 bounding box prediction is fed to a region of interest extractor and has the effect of superior localized bounding boxes that can contribute to enhanced feature extraction. Likewise, Cascade Mask R-CNN applies this concept to enhance mask predictions. Hybrid Task Cascade (HTC) introduces direct connections between the different stages of mask branches, thereby strengthening the flow by integrating complementary features across stages \cite{chen2019hybrid}.


To summarize, the most significant enhancements to instance segmentation in the literature include interventions at the level of the cascade, loss function, or backbone. In this paper, we propose to intervene at the level of the last prediction layer, in a manner to mitigate the problem of competing gradients between different classes. As far as we know, none of the previous methods do this. 

The contributions of this paper include the following:  
\begin{itemize}
\item Proposing a  modular Switch-Split block that can replace multi-class prediction heads in most instance segmentation methods. 
\item Ensuring no competing kernels between the different classes, which leads to richer representations as no trade-off between classes is permitted.
\item Enhancing the accuracy of instance segmentation models on the standard COCO dataset benchmark.
\end{itemize}

\section{Proposed System} 
\label{sec:methodology}
\begin{figure*}[t]
    \centering
    \includegraphics[width=6.7 in,height=2.8 in]{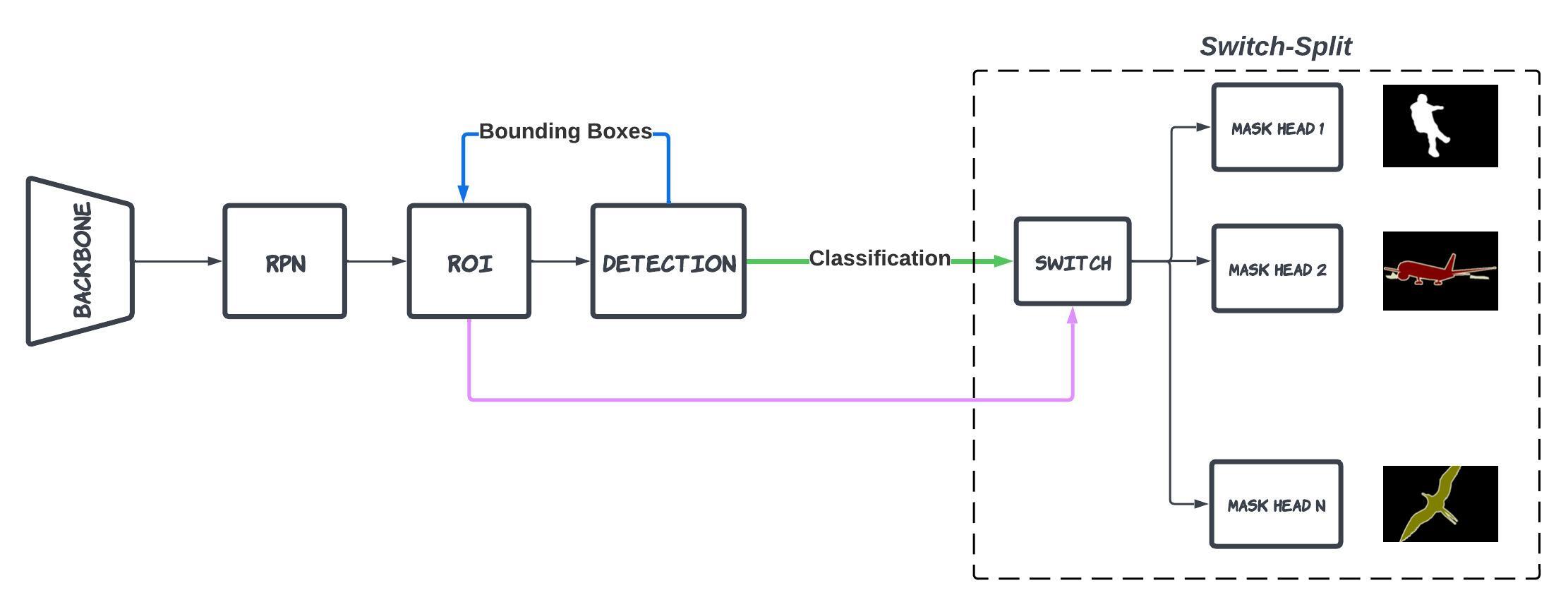}
    \caption{Switch-Split block architecture}
    \label{fig:ssblck}
\end{figure*}
Recalling from the previous section, at the output of an instance segmentation model, detected regions of interest are streamed into a terminal block consisting of three different branches: the first one is usually a classifier indicating the presence of an object in this region, and if so classifies it. The second branch predicts refined bounding box coordinates for each proposed region, adjusting the initially generated region proposals. The third branch takes the ROI-aligned features as input and predicts a binary mask for each object instance within the proposed regions. These masks indicate the pixel-level segmentation of the objects.

To our knowledge, all instance segmentation architectures in the literature include the aforementioned terminal block (\textit{i.e.}, including the three branches).  An interesting observation we came to realize is that this block in its current form suffers from what we refer to as competing kernels. The competing kernels problem arises when a model tries to \textit{synchronously} learn multiple classes having different features. During training, the kernels learned for each class compete with others from another class in an attempt to maximize their own segmentation accuracy. 

The main intuition driving our proposed MaskUno system is that each class should be learned on its own in a block we name the Switch-Split block.

\subsection{Switch-Split block}
The idea behind adding a switch and split block is to first benefit from the bounding box refinement and then specialize the learning per class. In other words, we first pass the bounding box head output through the ROI Align layer, then use it as an input to the Switch-Split block (see Fig.\ref{fig:ssblck}). The output of the classifier is given as an input to the switch, then the classifier, based on the input, classifies the refined ROI which then turns the switch to the corresponding class inside the splitting block. Finally, each ROI is assigned to a specific one-class mask head. 

The split block is a set containing $N+1$ blocks, each trained to classify and predict the bounding box and the mask of one specific class.

This idea applies to any architecture based on Mask-RCNN, and to any class. In this section, we will show how to adapt it to the baseline Mask-RCNN, the Cascade Mask-RCNN, and the hybrid task cascade. 

\subsubsection{Mask-RCNN}
instead of having the bounding boxes and the masks predicted in parallel as is usually done in Mask-RCNN, in MaskUno it is done sequentially. In other words, the bounding boxes are first predicted, and then based on these refined boxes the mask is estimated. Figure \ref{fig:ssblck} presents an example of the methodology applied to this architecture.

\begin{figure}[ht]
    \centering
    \includegraphics[scale=0.42]{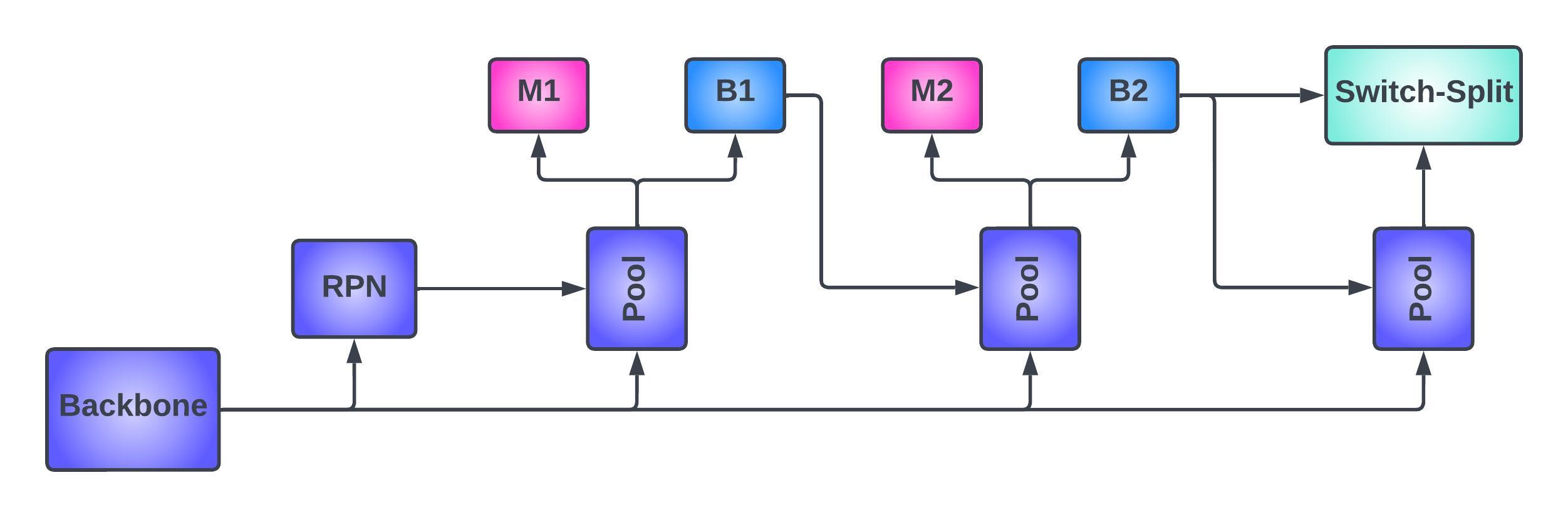}
    \caption{Switch-Split applied to Cascade Mask-RCNN}
    \label{fig:cascadessblck}
\end{figure}
\begin{figure}[ht]
    \centering
    \includegraphics[scale=0.42]{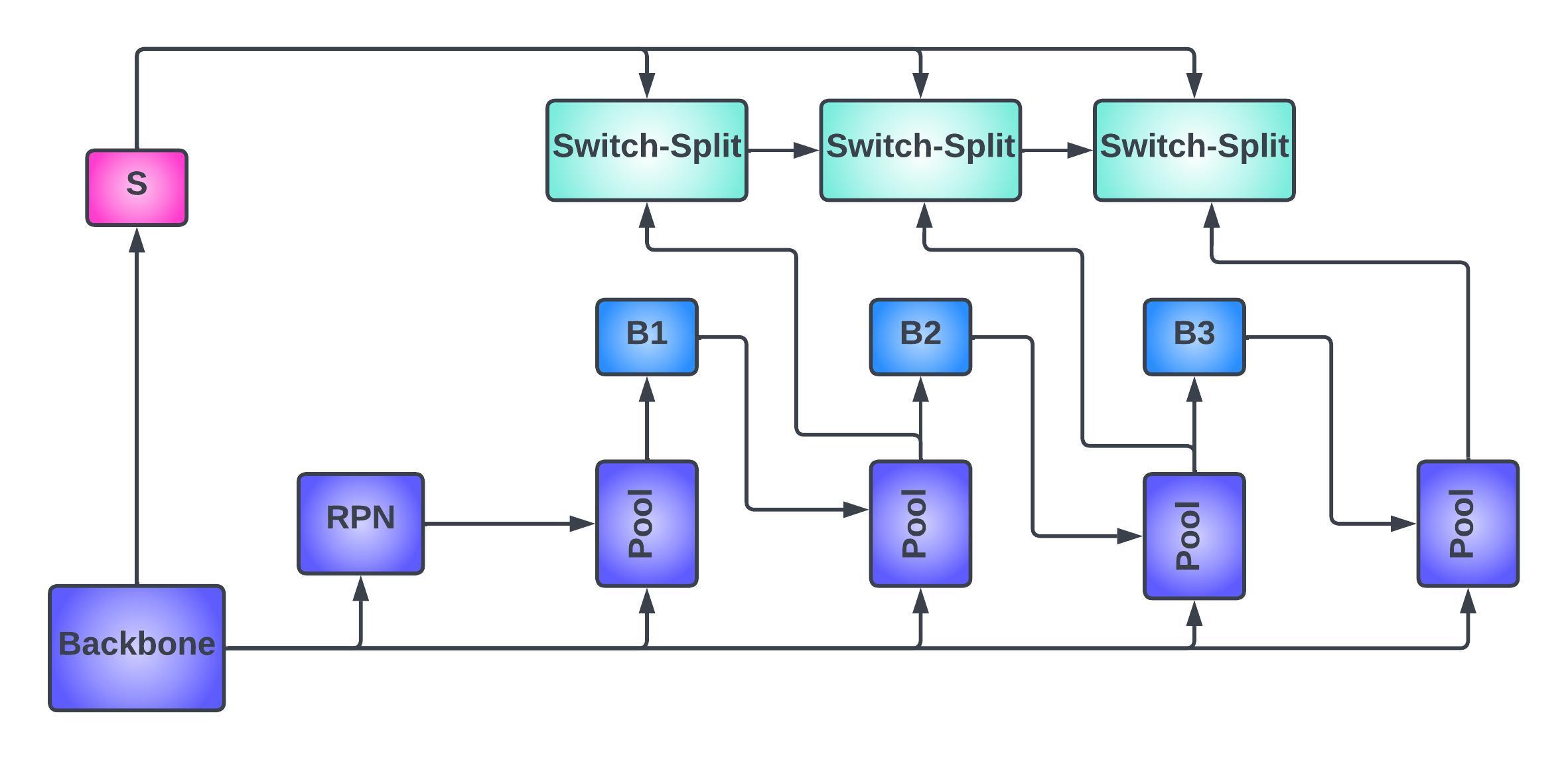}
    \caption{Switch-Split applied to Hybrid Task Cascade}
    \label{fig:htcssblck}
\end{figure}

\subsubsection{Cascade Mask-RCNN and Hybrid Task Cascade}

this architecture already includes bounding box refinements from one layer to the other, meaning that the output of the refined bounding box of Layer $i$+1, is fed as an input to the Switch-Split block of Layer $i$.
The split block is a set containing $N+1$ blocks, each trained to classify, and predict the mask of one specific class. Figure \ref{fig:cascadessblck} shows how each block is represented: they all include a mask head, all having the same functionality as the branches of the basic Mask-RCNN pipeline. In terms of hybrid-task cascade, we also replace every mask-head with a Switch-Split block, taking as input the refined classified bounding boxes as shown in Fig. \ref{fig:htcssblck}.

\subsection{Loss functions}

Every one of these architectures is allocated a set of three loss functions, one for every sub-part of this block. Hence, there is the categorical cross-entropy loss defined by (\ref{eq::CE}), where $y_i$ is the true class label, and $\hat{y}_i$ is the predicted class label, and $N$ is the number of classes. 

\begin{equation}
    L_{cls} = - \sum_{i=1}^N y_i \log(\hat{y}_i)
\label{eq::CE}
\end{equation}

The second loss function is the smooth L1 loss defined by (\ref{eq::L1loss}) where $x$ represents the difference between the predicted and the true value.

\begin{equation}
    \text{L1}(x) = \begin{cases}
0.5x^2, & \text{if } |x| < 1 \\
|x| - 0.5, & \text{otherwise}
\end{cases}
\label{eq::L1loss}
\end{equation}

For the mask prediction part, the loss function is a pixel-wise binary cross entropy defined by (\ref{eq::MaskLoss}).

\begin{equation}
    \begin{aligned}
        L_{mask} = - \sum_{i,j} [ y_{\text{true}}(i, j) \log(y_{\text{pred}}(i, j)) + 
        \\(1 - y_{\text{true}}(i, j)) \log(1 - y_{\text{pred}}(i, j)) 
    \end{aligned}
\label{eq::MaskLoss}
\end{equation}
where $y_{true}$ is the true pixel label, and $y_{pred}$ is the predicted pixel label. 

\subsection{Training}
A typical model contains its own set of loss functions [$L_{cls},  L1,  L_{mask}$] that it will try to minimize, while eliminating the issue of competing kernels. The mask head of class $i$ will have its own $L_{mask_i}$ representing the pixel-wise loss function that will be minimized. Having different loss functions for every class is also essential to solve the issue of competing kernels as it makes the weights of every mask head independent of one another also in terms of cost.

The training schedule and hyper-parameters are kept the same for every block to avoid overfitting on a specific class. 

\subsection{Generalizing to other pipelines for instance segmentation}
The switch split method should be applicable to any architecture, by first identifying the blocks where classification, bounding box regression, and mask prediction, are implemented, and then splitting the segmentation block into multiple ones, each specialized for the concerned class.

As a proof of concept, one can even, perform sequential training for every architecture, meaning train the model on Class 1 to obtain a specialized mask-head for this class. Then, train a new mask-head using the original checkpoint of the imported model on Class 2, up until all needed classes have a specialized mask-head. 

In Section IV, we perform experiments on models in the literature to validate our proposed method.

\begin{table*}[ht]
\centering
\begin{tabular}{llll|ll|ll|ll|ll|ll}
\cline{2-14}
 &
  Metric &
  \multicolumn{2}{l}{mAP} & 
  \multicolumn{2}{l}{mAP 50} &
  \multicolumn{2}{l}{mAP 75} &
  \multicolumn{2}{l}{mAP small} &
  \multicolumn{2}{l}{mAP medium} &
  \multicolumn{2}{l}{mAP large} \\ \hline
\multicolumn{1}{l|}{Class} &
   &
  Before &
  After &
  Before &
  After &
  Before &
  After &
  Before &
  After &
  Before &
  After &
  Before &
  After \\ \hline
\multicolumn{1}{l|}{\multirow{4}{*}{Dog}}   & MASK      & 0.58     & \textbf{0.622}     & 0.835     & \textbf{0.894}     & 0.7     & \textbf{0.759}     & 0.332     & \textbf{0.426}     & 0.561     & \textbf{0.601}     & 0.631     & \textbf{0.665}     \\
\multicolumn{1}{l|}{}                       & CASCADE   & 0.618     & \textbf{0.667}    & 0.825     & \textbf{0.893}     & 0.743    & \textbf{0.8}    & 0.347     & \textbf{0.456}     & 0.592     & \textbf{0.647}     & 0.671     & \textbf{0.708} \\
\multicolumn{1}{l|}{}                       & HTC       & 0.654     & \textbf{0.673}     & 0.878     & \textbf{0.901}     & 0.789     & \textbf{0.819}     & 0.43     & \textbf{0.462}     & 0.623    & \textbf{0.638}     & 0.706     & \textbf{0.725}     \\
\multicolumn{1}{l|}{}                       & DETECTORS & 0.669 & \textbf{0.687} & 0.887 & \textbf{0.906} & 0.813 & \textbf{0.834} & 0.4   & \textbf{0.454} & 0.641 & \textbf{0.659} & 0.724 & \textbf{0.732} \\ \hline

\multicolumn{1}{l|}{\multirow{4}{*}{Cup}}   & MASK      & 0.43     & \textbf{0.451}     & 0.634     & \textbf{0.666}     & 0.479     & \textbf{0.499}     & 0.263     & \textbf{0.284}     & 0.555     & \textbf{0.575}    & 0.708     & \textbf{0.722}     \\

\multicolumn{1}{l|}{}                       & CASCADE   & 0.457     & \textbf{0.514}     & 0.641     & \textbf{0.731}     & 0.507     & \textbf{0.573}    & 0.28     & \textbf{0.341}     & 0.599     & \textbf{0.64}     & 0.678     & \textbf{0.761}     \\
\multicolumn{1}{l|}{}                       & HTC       & 0.502     & \textbf{0.516}     & 0.715     & \textbf{0.732}     & 0.552     & \textbf{0.57}     & 0.336     & \textbf{0.348}     & 0.627     & \textbf{0.643}     & 0.75     & \textbf{0.762}     \\
\multicolumn{1}{l|}{}                       & DETECTORS & 0.523 & \textbf{0.537} & 0.731 & \textbf{0.75}  & 0.58  & \textbf{0.603} & 0.346 & \textbf{0.36}  & 0.656 & \textbf{0.665} & 0.794 & \textbf{0.8}   \\ \hline
\multicolumn{1}{l|}{\multirow{4}{*}{Cat}} &
  MASK &
  0.593 &
  \textbf{0.643} &
  0.911 &
  \textbf{0.94} &
  0.646 &
  \textbf{0.725} &
  0.18 &
  \textbf{0.389} &
  0.586 &
  \textbf{0.637} &
  0.609 &
  \textbf{0.655} \\
\multicolumn{1}{l|}{}                       & CASCADE   & 0.699     & \textbf{0.712}     & 0.942     & \textbf{0.954}    & 0.806     & \textbf{0.827}    & 0.293     & \textbf{0.293}     & 0.628     & \textbf{0.68}     & 0.73.     & \textbf{0.734}     \\ 
\multicolumn{1}{l|}{}                       & HTC       & 0.717 & \textbf{0.735} & 0.954 & \textbf{0.969} & 0.842 & \textbf{0.858} & 0.446 & \textbf{0.577} & 0.645 & \textbf{0.689} & 0.743 & \textbf{0.753} \\
\multicolumn{1}{l|}{}                       & DETECTORS & 0.741 & \textbf{0.759} & 0.951 & \textbf{0.964} & 0.866 & \textbf{0.883} & 0.48  & \textbf{0.56}  & 0.643 & \textbf{0.667} & 0.771 & \textbf{0.785} \\ \hline

\multicolumn{1}{l|}{\multirow{4}{*}{Book}}  & MASK      & 0.099     & \textbf{0.106}     & 0.224     & \textbf{0.238}     & 0.072    & \textbf{0.087}     & 0.058     & \textbf{0.062}     & 0.183    & \textbf{0.206}    & \textbf{0.346}     & 0.33     \\
\multicolumn{1}{l|}{}                       & CASCADE   & 0.127     & \textbf{0.15}     & 0.256     & \textbf{0.303}     & 0.114     & \textbf{0.136}     & 0.07     & \textbf{0.084}     & 0.236     & \textbf{0.287}    & 0.402     & \textbf{0.462}    \\ 
\multicolumn{1}{l|}{}                       & HTC       & 0.152    & \textbf{0.159}     & 0.312     & \textbf{0.319}     & 0.134     & \textbf{0.143}    & 0.092     & \textbf{0.095}     & 0.271     & \textbf{0.288}    & 0.474     & \textbf{0.484}     \\
\multicolumn{1}{l|}{}                       & DETECTORS & 0.173 & \textbf{0.183} & 0.35  & \textbf{0.363} & 0.155 & \textbf{0.168} & 0.11  & \textbf{0.114} & 0.313 & \textbf{0.335} & 0.504 & \textbf{0.517} \\ \hline

\multicolumn{1}{l|}{\multirow{4}{*}{Train}}     & MASK      & 0.603     & \textbf{0.62}     & 0.83     & \textbf{0.857}    & 0.677     & \textbf{0.697}     & 0.267    & \textbf{0.366}     & 0.23     & \textbf{0.264}     & 0.649     & \textbf{0.664}     \\
\multicolumn{1}{l|}{}                       & CASCADE & 0.653     & \textbf{0.675}    & 0.852     & \textbf{0.886}     & 0.73     & \textbf{0.743}     & \textbf{0.397}     & 0.347     & 0.26     & \textbf{0.267}    & 0.702     & \textbf{0.723}        \\ 
\multicolumn{1}{l|}{}                       & HTC & 0.676    & \textbf{0.692}     & 0.903     & \textbf{0.908}     & 0.739     & \textbf{0.759}     & 0.354     & \textbf{0.38}    & \textbf{0.316}     & 0.313     & 0.727     & \textbf{0.74}     \\
\multicolumn{1}{l|}{}                       & DETECTORS       & 0.71     & \textbf{0.713}    & 0.927     & \textbf{0.932}     & 0.789     & \textbf{0.793}     & \textbf{0.396}     & 0.391     & 0.284     & \textbf{0.314}     & \textbf{0.763}     & 0.758    \\ \hline

\multicolumn{1}{l|}{\multirow{4}{*}{TV}}    & MASK      & 0.574     & \textbf{0.598}     & 0.779     & \textbf{0.809}    & 0.68     & \textbf{0.704}     & 0.122    & \textbf{0.156}     & 0.532     & \textbf{0.552}     & 0.725     & \textbf{0.739}     \\
\multicolumn{1}{l|}{}                       & CASCADE   & 0.633     & \textbf{0.663}     & 0.81     & \textbf{0.856}     & 0.757     & \textbf{0.795}     & 0.182     & \textbf{0.234}     & 0.596     & \textbf{0.633}     & 0.772    & \textbf{0.794}     \\ 
\multicolumn{1}{l|}{}                       & HTC       & 0.657     & \textbf{0.664}     & 0.853    & \textbf{0.871}     & 0.763     & \textbf{0.775}     & 0.214     & \textbf{0.223}     & 0.623     & \textbf{0.634}     & 0.792     & \textbf{0.794}     \\
\multicolumn{1}{l|}{}                       & DETECTORS & 0.68  & \textbf{0.695} & 0.876 & \textbf{0.886} & 0.779 & \textbf{0.807} & \textbf{0.307} & 0.249 & 0.645 & \textbf{0.667} & 0.793 & \textbf{0.814} \\ \hline

\multicolumn{1}{l|}{\multirow{4}{*}{Bus}}   & MASK      & 0.622     & \textbf{0.648}     & 0.816     & \textbf{0.846}    & 0.689    & \textbf{0.721}     & 0.259    & \textbf{0.265}     & 0.451     & \textbf{0.49}     & 0.776     & \textbf{0.789}     \\
\multicolumn{1}{l|}{}                       & CASCADE   & 0.661     & \textbf{0.691}     & 0.844     & \textbf{0.885}     & 0.743     & \textbf{0.771}     & 0.293     & \textbf{0.329}     & 0.492     & \textbf{0.549}     & 0.804     & \textbf{0.819}    \\
\multicolumn{1}{l|}{}                       & HTC       & 0.701     & \textbf{0.711}     & 0.879     & \textbf{0.893}    & 0.785     & \textbf{0.809}     & 0.293     & \textbf{0.302}     & 0.568     & \textbf{0.587}     & 0.838     & \textbf{0.839}     \\
\multicolumn{1}{l|}{}                       & DETECTORS & 0.716 & \textbf{0.72}  & 0.9   & \textbf{0.903} & 0.821 & \textbf{0.823} & 0.326 & 0.316 & 0.589 & \textbf{0.596} & 0.843 & 0.843 \\ \hline

\multicolumn{1}{l|}{\multirow{4}{*}{Bird}}  & MASK      & 0.296     & \textbf{0.309}     & 0.558     & \textbf{0.584}    & 0.274     & \textbf{0.288}     & 0.183    & \textbf{0.194}     & 0.481     & \textbf{0.499}     & 0.637     & \textbf{0.656}     \\
\multicolumn{1}{l|}{}                       & CASCADE   & 0.317     & \textbf{0.335}     & 0.56     & \textbf{0.611}     & 0.31     & \textbf{0.323}     & 0.187     & \textbf{0.202}     & 0.517     & \textbf{0.542}     & 0.699     & \textbf{0.711}    \\ 
\multicolumn{1}{l|}{}                       & HTC       & 0.339     & \textbf{0.342}     & 0.601     & \text{0.608}    & 0.348     & \textbf{0.35}     & 0.201     & \textbf{0.207}     & 0.547     & \textbf{0.551}     & 0.738     & 0.732 \\
\multicolumn{1}{l|}{}                       & DETECTORS & 0.343 & \textbf{0.352} & 0.604 & \textbf{0.614} & \textbf{0.337} & 0.334 & 0.207 & \textbf{0.213} & 0.552 & \textbf{0.565} & 0.75  & \textbf{0.768} \\ \hline

\multicolumn{1}{l|}{\multirow{4}{*}{Cow}}   & MASK      & 0.461     & \textbf{0.48}     & 0.753     & \textbf{0.794}    & 0.512     & \textbf{0.529}     & 0.34    & \textbf{0.361}     & 0.5     & \textbf{0.519}     & 0.611     & \textbf{0.626}     \\
\multicolumn{1}{l|}{}                       & CASCADE   & 0.516     & \textbf{0.549}     & 0.794     & \textbf{0.845}     & 0.595     & \textbf{0.635}     & 0.388     & \textbf{0.427}     & 0.545     & \textbf{0.577}     & 0.68     & \textbf{0.708}    \\ 
\multicolumn{1}{l|}{}                       & HTC       & 0.533     & \textbf{0.548}     & 0.837     & \textbf{0.852}    & 0.613     & \textbf{0.634}     & 0.418     & \textbf{0.429}     & 0.554     & \textbf{0.57}     & 0.694     & \textbf{0.705}     \\
\multicolumn{1}{l|}{}                       & DETECTORS & 0.561 & \textbf{0.569} & 0.867 & \textbf{0.871} & 0.654 & \textbf{0.67}  & 0.426 & 0.426 & 0.579 & \textbf{0.595} & 0.744 & \textbf{0.749} \\ \hline

\multicolumn{1}{l|}{\multirow{4}{*}{Sheep}} & MASK      & 0.422     & \textbf{0.435}     & 0.711     & \textbf{0.735}    & 0.456     & \textbf{0.472}     & 0.177    & \textbf{0.197}     & 0.491     & \textbf{0.503}     & 0.562     & \textbf{0.575}     \\
\multicolumn{1}{l|}{}                       & CASCADE   & 0.499     & \textbf{0.508}     & 0.786     & \textbf{0.799}     & 0.54     & \textbf{0.549}     & 0.242     & \textbf{0.253}     & 0.566     & \textbf{0.569}     & 0.65     & \textbf{0.662}    \\ 
\multicolumn{1}{l|}{}                       & HTC       & 0.494     & \textbf{0.504}     & 0.793     & \textbf{0.804}    & 0.526     & \textbf{0.537}     & 0.216     & \textbf{0.23}     & 0.554     & \textbf{0.564}     & 0.681     & \textbf{0.687}     \\
\multicolumn{1}{l|}{}                       & DETECTORS & 0.534 & \textbf{0.539} & 0.836 & \textbf{0.838} & 0.579 & \textbf{0.581} & 0.308 & \textbf{0.311} & 0.58  & \textbf{0.583} & 0.71  & \textbf{0.714} \\ \hline
\end{tabular}
\caption{Comparative results of all models on different metrics for 10 classes}
\label{tab:table1}
\end{table*}

\section{Experimentation and Discussion}
\label{sec:Experimentation and Discussion}




\subsection{Datasets and metrics}
\subsubsection{Data used}
Experiments are performed on the Common Object in Context (COCO) dataset \cite{cocodataset} because it is challenging due to the numerous classes it contains, and its high number of samples, but also to make a fair comparison with the state-of-the-art models trained using this benchmark. We use the train2017 and val2017 data which contain a total of 115,000 and 5,000 images respectively. The data is split per-class for both training and validation. For the bounding box and mask training, instance annotations are used, and for the semantics, the COCO-stuff annotations.


\subsubsection{Evaluation}

to evaluate the model performance, we use the mean average precision taken at different IOU thresholds (at 0.5 and 0.75). Additionally, different scales are taken for the data: small, medium, and large.  We report the AP, $AP_{50}$, $AP_{75}$, $AP_s$, $AP_m$, and $AP_l$.
\subsection{Experimentation}

To validate the generalization of MaskUno we test it on various classes and various architectures. The process starts by removing the multiclass mask prediction head from a pre-trained model with a plateau in its accuracy, followed by adding the Switch-Split block and finally training the new weights on the sought classes. Afterward, we compare its mean Average Precision (mAP) with the one obtained before training the model. The hypothesis is that its mean average precision increases for every class using the MaskUno method.
The experimentats are divided accordingly. First, using the COCO dataset, and the library MMDetection \cite{mmdetection}, a similar experiment is conducted on four different models: the baseline Mask-RCNN \cite{he2017mask}, the Cascade-Mask-RCNN \cite{cai2018cascade}, the Hybrid Task Cascade \cite{chen2019hybrid}, and the DetectoRS backbone \cite{qiao2021detectors} on a list of 10 classes. Since the pre-trained models have more classes compared to the MaskUno models, we separate the validation data into different sub-datasets, each responsible for validating a specific class. 
Furthermore, we experiment on a total of 80 classes and compare the results for Mask-RCNN and DetectoRS. Mask-RCNN is used since it is the baseline model for instance segmentation. DetectoRS is an architecture that contains a change in backbone, as well as a hybrid task cascade architecture. 
Note that, given the methodology, we care about the mask prediction results for all models and classes.

\subsection{Results and discussion}

We discuss in this section the results obtained after applying our method to different models and a variety of classes. Table \ref{tab:table1} shows the results of the four models on every class. The experiments consist of training each of these previously assumed saturated models, on a list of ten classes, shown in the column of the table. The models are chosen in a way to include the different techniques found in the literature, including (1) a change in the backbone (DetectoRS), (2) a cascade architecture (Cascade mask-RCNN), and (3) a hybrid task (HTC). It is worth mentioning that DetectoRS is currently the best-performing CNN-based model for instance segmentation. 
Moreover, we can see comparative results of the metrics before and after training, and for all architectures, there is a significant increase in the mAP which varies from 0.5\% to 5\%.   

Furthermore, for some classes, we did not observe a significant increase in the metrics. A possible explanation would be the number of samples available. One example is the class 'sheep' which contains a small number of samples (765 samples for training). Additionally, very few entries show a slight decrease after applying our method.

\begin{figure}
    \centering
    \includegraphics[scale=0.6]{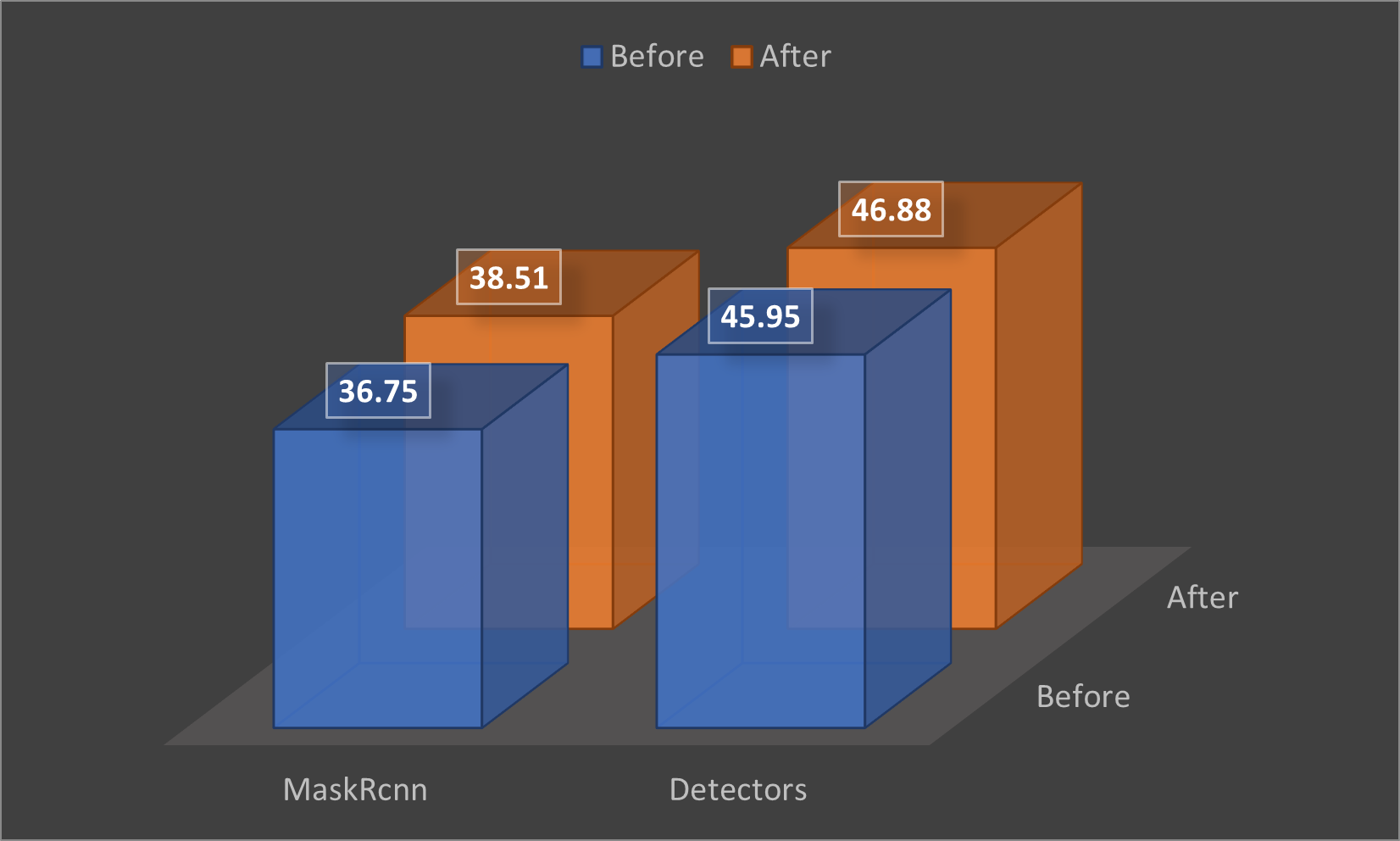}
    \caption{Bar-graph showing the increase in the mAP before and after using MaskUno for Mask-RCNN and DetectoRS}
    \label{fig:maskunobar}
\end{figure}

\begin{figure*}[h]
    \centering
    \includegraphics[width=7.2in,height=6.9 in]{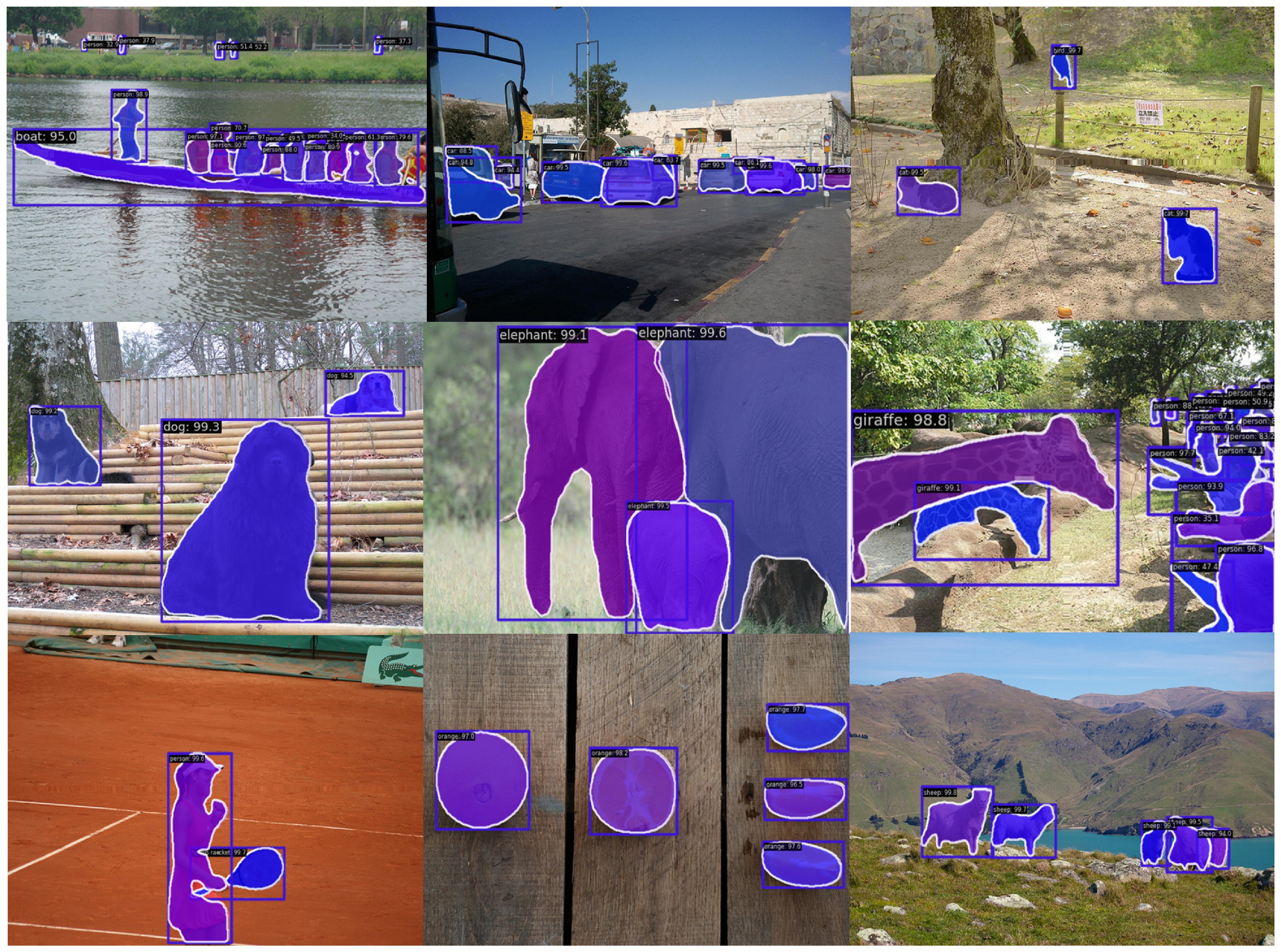}
    \caption{Examples of various predictions}
    \label{fig:examples}
\end{figure*}

This experiment supports the fact that MaskUno can be applied to a wide set of methods for instance segmentation. 
We extend this experiment by training 80 classes for both the baseline Mask-RCNN and DetectoRS. For both models, we observe an average increase in the mAP (Fig.\ref{fig:maskunobar}).  This enhancement infers a different conclusion for every model. For Mask-RCNN, the mAP increases by 4.8\%, which confirms that MaskUno is not affected by a specific choice of class. As for DetectoRS, the mAP increases by 2.03\%, which validates that solving the competing kernels is complementary to a cascade architecture or a backbone enhancement. 
Figure \ref{fig:examples} shows a few examples at inference time, for multiple classes. Besides successful detection and segmentation, one can note that the degree of confidence for every instance is significantly high.




\section{Limitations and Future Work}
\subsubsection*{Limitations}
Our experiments demonstrate that MaskUno outperforms previous models when evaluated on a per-class basis, with each class evaluated exclusively on its positive images. However, it's important to note that this experimental setup does not account for negative frames where false positives may cause the switch to misdirect certain images to incorrect branches.
\subsubsection*{Further experiments}

Another experiment would be to investigate the effect of splitting the bounding box regression block on the mAP, compare it to classical bounding box refinement, and merge both together. A new version of MaskUno where splitting is done on the branch responsible for bounding box prediction, followed by the mask split. This would not only enhance bounding box refinements but also the segmentation branch benefiting from this refinement. 

\subsubsection*{Transformer-based models}

Instance segmentation models based on transformers proved to surpass CNN-based models. They are usually used as a backbone for various computer vision downstream tasks. A similar approach could be implemented, where the bounding box regression and segmentation branches of such models would be split into multiple parts to specialize the training per-class. 

\section{Conclusion}
\label{sec: conclusion}
To address the issue of conflicting gradients, we offer MaskUno, a Switch-Split block that can be used for most instance segmentation algorithms. It enhances the accuracy of instance segmentation models based on Mask-RCNN that use multi-class segmentation methods and has a complementary effect to tasks such as cascade or hybrid task. MaskUno provided a 2.03\% improvement over one of the best-performing CNN-based instance segmentation models "Detectors" on the standard COCO dataset. Moreover, for future work, applying MaskUno on transformer-based models, while benefiting from a split of both bounding box regression and mask prediction, has the potential to reach a new state-of-the-art accuracy.

\section*{Acknowledgment}
This work was supported by the Didymos Horizon Europe project, grant number 101092875–DIDYMOS-XR.

\bibliographystyle{ieeetr}
\bibliography{ref}
\end{document}